\pdfoutput=1

\documentclass[11pt]{article}

\usepackage{acl}

\usepackage{times}
\usepackage{latexsym}
\usepackage{paralist}
\usepackage{booktabs}
\usepackage{graphicx}
\usepackage{float}

\usepackage[T1]{fontenc}

\usepackage[utf8]{inputenc}

\usepackage{microtype}

\title{Sub-Word Alignment is Still Useful: A Vest-Pocket Method for Enhancing Low-Resource Machine Translation}

\author{Minhan Xu, Yu Hong\thanks{\quad Corresponding author.} \\
  School of Computer Science and Technology, Soochow University, China \\
  \texttt{cosmosbreak5712@gmail.com, tianxianer@gmail.com}}

\begin{document}
\maketitle
\begin{abstract}
We leverage embedding duplication between aligned sub-words to extend the Parent-Child transfer learning method, so as to improve low-resource machine translation. We conduct experiments on benchmark datasets of My$\rightarrow$En, Id$\rightarrow$En and Tr$\rightarrow$En translation scenarios. The test results show that our method produces substantial improvements, achieving the BLEU scores of 22.5, 28.0 and 18.1 respectively. In addition, the method is computationally efficient which reduces the consumption of training time by 63.8\%, reaching the duration of 1.6 hours when training on a Tesla 16GB P100 GPU. All the models and source codes in the experiments will be made publicly available to support reproducible research.
\end{abstract}

\section{Introduction}
Low-resource machine translation (MT) is challenging due to the scarcity of parallel data and, in some cases, the absence of bilingual dictionaries \citep{zoph-etal-2016-transfer,miceli-barone-2016-towards,koehn-knowles-2017-six,zhang-etal-2017-adversarial}. Unsupervised, multilingual and transfer learning have been proven effective in the low-resource MT tasks, grounded on different advantages (section 2).   

In this paper, we follow \citet{aji-etal-2020-neural}'s work to utilize cross-language transfer learning, of which the ``parent-child'' transfer framework is first proposed by \citet{zoph-etal-2016-transfer}. In the parent-child scenario, a parent MT model and a child MT model are formed successively, using the same neural network structure. In order to achieve the sufficient warm-up effect from scratch, the \textbf{parent} is trained on \textbf{high}-resource language pairs. Further, the \textbf{child} inherits the parent's properties (e.g., inner parameters and embedding layers), and it is boosted by the fine-tuning over \textbf{low}-resource language pairs. One of the distinctive contributions in \citet{aji-etal-2020-neural}'s study is to demonstrate the significant effect of embedding duplication for transference, when it is conducted between the morphologically-identical sub-words in different languages.

We attempt to extend \citet{aji-etal-2020-neural}'s work by additionally duplicating embedding information among the aligned multilingual sub-words. It is motivated by the assumption that if the duplication between morphologically-identical sub-words contributes to cross-language transference, the duplication among any other type of equivalents is beneficial in the same way, such as that of the aligned sub-words, most of which are likely to be morphologically-dissimilar but semantically-similar (or even exactly the same).

In our experiments, both the parent and child MT models are built with the transformer-based \citep{vaswani2017attention} encoder-decoder architecture (Section 3.1). We use the unigram model from SentencePiece \citep{kudo-richardson-2018-sentencepiece} for tokenizing, and carry out sub-word alignment using eflomal (Section 3.2). On the basis, we develop a normalized element-wise embedding aggregation method to tackle the many-to-one embedding duplication for aligned sub-words (Section 3.3). The experiments show that our method achieves substantial improvements without using data augmentation.

\section{Related Work}
The majority of previous studies can be sorted into 3 aspects in terms of the exploited learning strategies, including unsupervised, multilingual and transfer learning.
\begin{itemize}
\item {\textbf{Unsupervised}} MT conducts translation merely conditioned on monolingual language models \citep{lample2018unsupervised,artetxe2017unsupervised}. The ingenious method that has been explored successfully is to bridge the source and target languages using a shareable representation space \citep{lample-etal-2018-phrase}, which is also known as interlingual \citep{cheng2019joint} or cross-language embedding space \citep{kim-etal-2018-improving}. To systematize unsupervised MT, most (although not all) of the arts leverage bilingual dictionary induction \citep{conneau2018word,sogaard-etal-2018-limitations}, iterative back-translation \citep{sennrich-etal-2016-improving,lample-etal-2018-phrase} and denoised auto-encoding \citep{vincentextracting,kim-etal-2018-improving}. 
\item {\textbf{Multilingual}} MT conducts translation merely using a single neural model whose parameters are thoroughly shared by multiple language pairs \citep{firat-etal-2016-multi,lee-etal-2017-fully,johnson-etal-2017-googles,gu-etal-2018-universal,gu-etal-2018-meta}, including a variety of high-resource language pairs as well as a kind of low-resource (the target language is fixed and definite). Training on a mix of high-resource and low-resource (even zero-resource) language pairs enables the shareable model to generalize across language boundaries \citep{johnson-etal-2017-googles}. The benefits result from the assimilation of relatively extensive translation experience and sophisticated modes from high-resource language pairs.
\item {\textbf{Transferable}} MT is fundamentally similar to multilingual MT, whereas it tends to play the aforementioned Parent-Child game \citep{zoph-etal-2016-transfer}. A variety of optimization methods have been proposed, including the transfer learning over the embeddings of WordPieces tokens \citep{johnson-etal-2017-googles}, BPE sub-words \citep{nguyen-chiang-2017-transfer} and the shared multilingual vocabularies \citep{kocmi-bojar-2018-trivial,gheini2019universal}, as well as the transference that is based on the artificial or automatic selection of congeneric parent language pairs  \citep{dabre-etal-2017-empirical,lin-etal-2019-choosing}. In addition, \citet{aji-etal-2020-neural} verify the different effects of various transferring strategies of sub-word embeddings, such as that among morphologically-identical sub-words.
\end{itemize}

In this paper, we extend \citet{aji-etal-2020-neural}'s work, transferring embedding information not only among the morphologically-identical sub-words but the elaborately-aligned sub-words.

\section{Approach}

\subsection{Preliminary: Basic Transferable NMT}
We follow \citet{kim-etal-2019-effective} and \citet{aji-etal-2020-neural} to build neural MT (NMT) models with 12-layer transformers \citep{vaswani2017attention}, in which the first 6 layers are used as the encoder while the subsequent 6 layers the decoder.

\textbf{Embedding Layer} As usual, the encoder is coupled with a trainable embedding layer, which maintains a fixed bilingual vocabulary and trainable sub-word embeddings. Each embedding is specified as a 512-dimensional real-valued vector. 


\textbf{Parent-Child Transfer} We follow \citet{zoph-etal-2016-transfer} to conduct Parent-Child transfer learning. Specifically, we adopt an off-the-shelf transformer-based NMT\footnote{https://github.com/Helsinki-NLP/OPUS-MT-train/blob/master/models/de-en/README.md} which was adequately trained on high-resource De$\rightarrow$En (German$\rightarrow$English) language pairs. The publicly-available data of OPUS \citep{tiedemann-2012-parallel} is used for training, which comprises about 351.7M De$\rightarrow$En parallel sentence pairs\footnote{https://opus.nlpl.eu/}. We regard this NMT model as the Parent. Further, we transfer all inner parameters of the 12-layer transformers from Parent to Child.

By contrast, the embedding layer of Parent is partially transferred to Child, which has been proven effective in \citet{aji-etal-2020-neural}'s study. Assume $V_h$ denotes the high-resource (e.g., the aforementioned De-En) vocabulary while $V_l$ the low-resource, the morphologically-identical sub-words $V_o$ are then specified as the ones occurring in both $V_h$ and $V_l$ (i.e., $V_o=V_h\cap V_l$). Thus, we duplicate the embeddings of morphologically-identical sub-words $V_o$ from the embedding layer of Parent to that of Child. Further, we randomly initialize the embeddings of the rest sub-words $V_r$ in the Child's embedding layer ($V_r=V_l-V_o$), where random sampling from a Gaussian distribution is used.

Both the transferred inner parameters and the duplicated embeddings constitutes the initial state of the Child NMT model. On the basis, we fine-tune Child on the low-resource language pairs, such as the considered 18K My$\rightarrow$En (Burmese$\rightarrow$English) parallel data in our experiments.

\begin{table}
\centering
\begin{tabular}{lccc} 
\hline
   & \textbf{Doc.} & \textbf{Sent.} & \textbf{Token}  \\ 
\hline
My & 113K & 1.1M      & 17.4M   \\
Id & 1.1M & 8.3M      & 156.2M  \\
Tr & 705K & 5.8M      & 128.2M  \\
\hline
\end{tabular}
\caption{Statistics of monolingual Wikipedia data.}
\end{table}

\subsection{Tokenizer and Alignment}
We strengthen Parent-Child transfer learning by additionally duplicating embeddings for aligned sub-words (between low and high-resource languages). The precondition is to 
produce the word-level alignment and equivalently assign it to sub-words.

\textbf{Word Alignment} We use Eflomal\footnote{https://github.com/robertostling/eflomal} to achieve the word alignment. It is developed based on EFMARAL \citep{ostling2016efficient}, where Gibbs sampling is run for inference on Bayesian HMM models. Eflomal is not only computationally efficient but able to perform $n$-to-1 alignment. We separately train Eflomal on the low-resource My$\rightarrow$En, Id (Indonesian)$\rightarrow$En and Tr (Turkish)$\rightarrow$En parallel data (Section 4).

\textbf{Sub-word Tokenizer} We train a sub-word tokenizer using the unigram model of SentencePiece for each low-resource language, including My, Id and Tr. The tokenizers are trained on monolingual plain texts which are collected from Wikipedia's dumps\footnote{https://dumps.wikimedia.org}. The toolkit wikiextractor\footnote{https://github.com/attardi/wikiextractor} is utilized to extract plain texts from the semi-structured data. The statistics of training data is shown in Table 1.

We uniformly set the size of sub-word vocabulary to 50K when training the tokenizers. The obtained vocabulary of each low-resource language is utilized for sub-word alignment, towards the mixed De-En sub-word vocabulary in the Parent NMT model. The size of De-En vocabulary is 58K.

\textbf{Sub-word Alignment} Given a pair of aligned bilingual words, we construct the same correspondence for their sub-words by many-to-many mappings. See the De$\rightarrow$Tr example in (1).

\begin{itemize}
\item[(1)] Word Alignment: | {\em produktion}$\leftrightarrow${\em üretme}\\
\hphantom{Word Alignment:\,\,}| {\em Harnstoff}$\leftrightarrow${\em üre}\\
Sub-word Alignment:\,| {\em produck}$\leftrightarrow$\{{\em \textbf{üre}}, {\em tme}\}\\
\hphantom{Sub-word Alignment:\,}| {\em tion}$\leftrightarrow$\{{\em \textbf{üre}}, {\em tme}\}\\
\hphantom{Sub-word Alignment:\,}| {\em Harn}$\leftrightarrow$\{{\em \textbf{üre}}\}\\
\hphantom{Sub-word Alignment:\,}| {\em stoff}$\leftrightarrow$\{{\em \textbf{üre}}\}
\end{itemize}

It is unavoidable that some of the aligned sub-words are non-canonical. Though, the positive effect on transfer learning may be more substantial than negative. It motivated by the findings that the use of sub-words ensures a sufficient overlap between vocabularies \citep{nguyen-chiang-2017-transfer}, and thus enables the transfer of a larger number of concrete embeddings rather than random ones.

\begin{table}
\centering
\begin{tabular}{lccc}
\hline
      & \textbf{Train.} & \textbf{Val.} & \textbf{Test} \\
\hline
My-En (ALT) & 18K   & 1K  & 1K   \\
Id-En (BPPT) & 22K   & 1K  & 1K   \\
Tr-En (WMT17) & 207K  & 3K  & 3K   \\
\hline
\end{tabular}
\caption{Statistics for low-resource parallel datasets.}
\end{table}

\subsection{$N$-to-1 Embedding Duplication}
Assume that  $V_l^a$ denotes the sub-words in low-resource vocabulary that have aligned sub-words in high-resource vocabulary, the mapping is $D(x)$, note that $\forall x \in V_l^a$, $D(x)$ is a set of sub-words. Thus, in the embedding layer of Child, we extend the range of sub-words for embedding transfer, including both the identical sub-words $V_o$ and the aligned $V_l^a$. To enable the transfer, we tackle $n$-to-1 embedding duplication. It is because that, in a large number of cases, there is more than one high-resource sub-word corresponding to a single low-resource sub-word (see ``{\em üre}'' in (1)).


Given a sub-word $x$ in $V_l^a$ and the aligned sub-words $v_x$ in $D(x)$, we rank $v_x$ in terms of the frequency with which they were found to be aligned with $x$ in the parallel data. On the basis, we carry out two duplication methods as below.

\begin{itemize}
\item {\textbf{Top-1}} We take the top-$1$ sub-word $\check{x}$ from $v_x$, and perform element-wise embedding duplication from $\check{x}$ to $x$: $\forall i, E_i(\check{x})=E_i(x)$ ($i$ is the $i$-th dimension of embedding $E(*)$).
\item {\textbf{Mean}} We adopt all the sub-words in $v_x$, and duplicate their embedding information by the normalized element-wise aggregation (where, $n$ denotes the number of sub-words in $v_x$): $$\forall i, E_i(\check{x})=\sum_{x\in v_x}E_i(x)/n$$ 
\end{itemize}

\section{Experimentation}

\subsection{Datasets and Evaluation Metric}
We evaluate the transferable NMT models for three source languages (My, Id and Tr). English is invariably specified as the target language. There are three low-resource parallel datasets used for training the Child NMT model, including Asian
Language Treebank (ALT) \citep{ding2018nova}, PAN Localization BPPT\footnote{http://www.panl10n.net/english/OutputsIndonesia2.htm} and the corpus of WMT17 news translation task \citep{bojar-etal-2017-findings}. The statistics in the training, validation and test sets is shown in Table 2. We evaluate all the considered NMT models with SacreBLEU \citep{post2018call}.

\begin{table}
\centering
\begin{tabular}{lccc}
\hline
Model & \textbf{My-En} & \textbf{Id-En} & \textbf{Tr-En} \\ \hline
Baseline    &  20.5           & 26.0           & 17.0           \\
\hline
MI-PC    & 21.0           & 27.5           & 17.6           \\
Top-1-PC             & 21.9           & 27.6           & 18.0           \\
\textbf{Mean-PC}              & \textbf{22.5}           & \textbf{28.0}           & \textbf{18.1}           \\ \hline
\end{tabular}
\caption{Results using \underline{\textbf{SentencePiece}} tokenizer.}
\end{table}

\subsection{Hyperparameters}
We use an off-the-shelf NMT model as Parent (Section 3.1), whose state variables (i.e., hyperparameters and transformer parameters) and embedding layer are all set. On the contrary, the Child NMT model needs to be regulated from scratch.

When training and developing Child, we adopt the following hyperparameters. Each source language was tokenized using SentencePiece \citep{kudo-richardson-2018-sentencepiece} with 50k vocabulary size. Training was carried out with HuggingFace Transformers library \citep{wolf-etal-2020-transformers} using the Adam optimizer with 0.1 weight decay rate.
The maximum sentence length was set to 128 and the batch size to 64 sentences. The learning rate was set to 5e-5 and checkpoint frequency to 500 updates. For each model, we selected the checkpoint with the lowest perplexity on the validation set for testing.


\begin{table}
\centering
\begin{tabular}{lccc}
\hline
Model & \textbf{My-En} & \textbf{Id-En} & \textbf{Tr-En} \\ \hline
Baseline    & 20.2           & 24.5           & 16.5           \\
\hline
MI-PC    & 20.4           & 24.2           & 16.8           \\
Top-1-PC            & 21.2           & 26.9           & 16.9           \\
\textbf{Mean-PC}              & \textbf{21.9}           & \textbf{27.1}           & \textbf{16.9}           \\ \hline
\end{tabular}
\caption{Results using \underline{\textbf{BPE}} tokenizer.}
\end{table}

\section{Results and Analysis}
Table 3 shows the test results, where all the considered \underline{P}arent-\underline{C}hild transfer models are marked with ``PC'', and the baseline is the transformer-based NMT (Section 3.1) which is trained merely using low-resource parallel data (without transfer learning). MI-PC is the reproduced transfer model in terms of \citet{aji-etal-2020-neural}'s study, in which only the embedding transference of morphologically-identical sub-words is used. We report NMT performance when MI-PC is used to enhance the baseline, as well as that when our auxiliary transfer models (i.e., Top-1 and Mean in Section 3.3) are additionally adopted, separately.

It can be observed that, compared to MI-PC, both Top-1-PC and Mean-PC yield improvements for all the three low-resource MT scenarios. The most significant improvement occurs for My$\rightarrow$En MT, reaching up to 1.5 BLEU. Both the models generalize well across changes in the input sub-words. It can be illustrated in a separate experiment where the BPE \citep{sennrich-etal-2016-neural} tokenizer is used (instead of SentencePiece \citep{kudo-richardson-2018-sentencepiece}), and all the transfer models are run over the newly-aligned sub-words. As shown in Table 4, both Top-1-PC and Mean-PC still outperform MI-PC, yielding an improvement of 2.9 BLEU at best (for Id$\rightarrow$En MT).



\begin{figure}[t]
\centering
\includegraphics[width=0.41\textwidth]{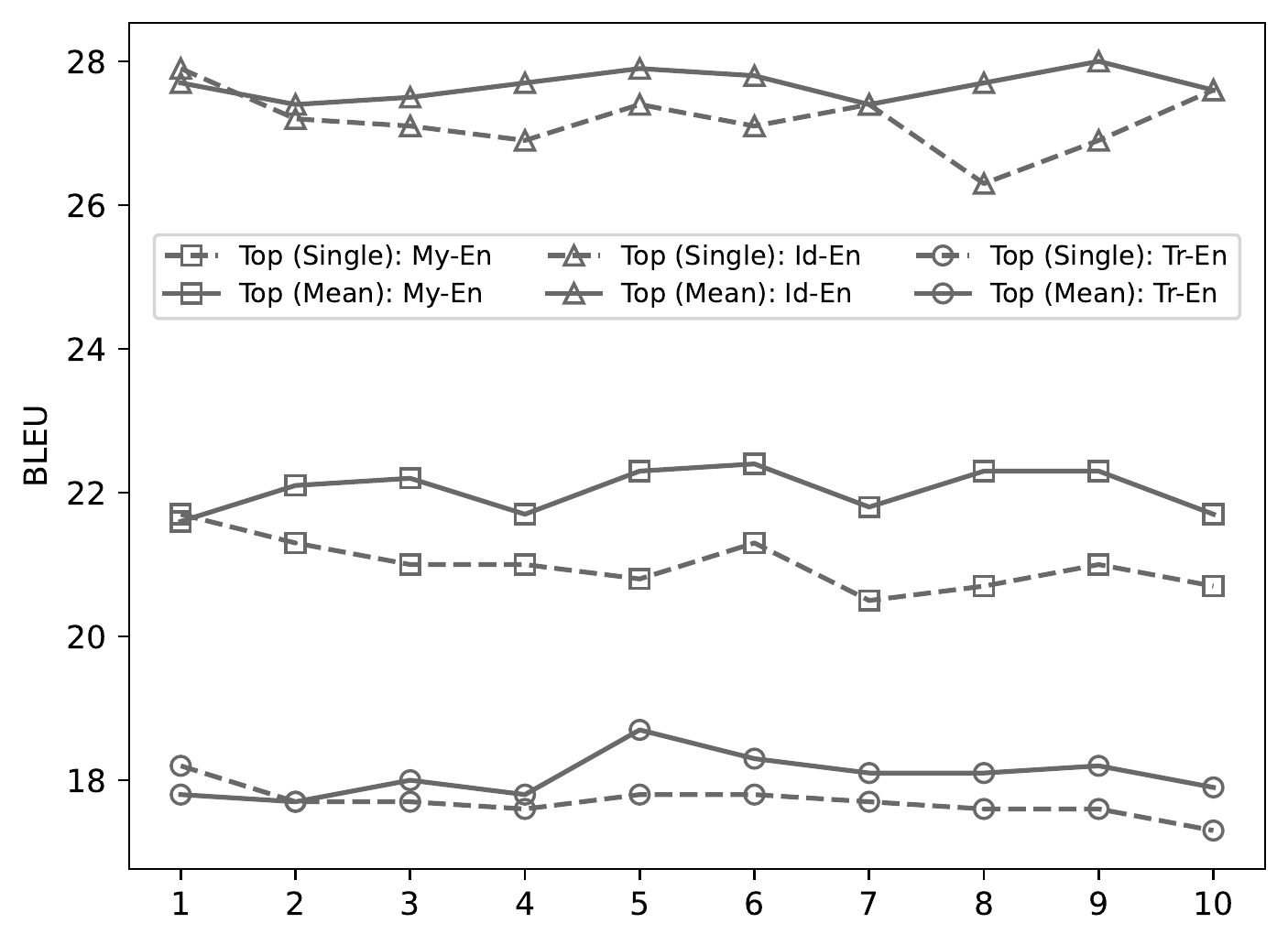}
\caption{Comparison between embedding duplication of a single aligned sub-word (denoted with \underline{\textbf{Single}}) and that of multiple sub-words (\underline{\textbf{Mean}}).}
\end{figure}

\begin{table}[t]
  \centering
  \begin{tabular}{lccc}
  \hline
  Model & \textbf{My-En} & \textbf{Id-En} & \textbf{Tr-En} \\ \hline
  Baseline    & 1.30           & 1.27           & 4.49           \\
  MI-PC    & 1.30           & 1.35           & 3.53           \\
  Top-1-PC            & 1.11           & 1.00           & 3.07           \\
  \textbf{Mean-PC}              & \textbf{0.96}           & \textbf{0.94}           & \textbf{2.14}           \\ \hline
  \end{tabular}
  \caption{The time (in hour) that different MT models consumed during training in all experiments (0.9 hour is equivalent to 54 minutes).}
  \label{tab:runtime}
\end{table}

Due to unavoidable errors in the sub-word alignment, the utilization of a single aligned sub-word for embedding duplication easily results in performance degradation.
Aggregating and normalizing embeddings of all possible aligned sub-words help to overcome the problem. Figure 1 shows the NMT performance obtained when the $i$-th top-ranked aligned sub-word is exclusively used for transfer, as well as the aggregation of top-$i$ sub-words is used. It can be found that the latter model almost always outperforms the former model.

We compare the training time consumption of all experiments, the result is shown in Table 5. We use mixed precision for training the child MT model. All experiments are conducted on a single NVIDIA P100 16GB GPU. 

Obviously, the time that Mean-PC consumes during training is less than other models. In the scenario of Tr-En MT, the training duration is even shortened from 4.49 hours (i.e., about 269 minutes) to 2.14, compared to the baseline model. Most probably, it is caused by the transferring of a larger number of sub-word embeddings during training. In other word, Mean-PC actually transfers not only morphologically-identical sub-words but the aligned ones. This contributes more to the avoidance of redundant learning over sub-word embeddings. All in all, Mean-PC is less time-consuming when producing substantial improvements.

\section{Conclusion}
We enhance transferable Parent-Child NMT by duplicating embeddings of aligned sub-words. The experimental results demonstrate that the proposed method yields substantial improvements for all the considered MT scenarios (including My-En, Id-En and Tr-En). More importantly, we successfully reduce the training duration. The efficiency can be improved with the ratio of about 50\% at best.

Additional survey in the experiments reveals that phonetic symbols can be used for transfer learning between the languages belonging to different families. For example, the phonologies of {\em hamburger} in German and Burmese are similar (H$\acute{a}$mburger vs hambhargar). In the future, we will study bilingual embedding transfer of phonologically-similar words, so as to further improve low-resource NMT.

\section*{Acknowledgements}
The research is supported by National Key R\&D Program of China (2020YFB1313601) and National Science Foundation of China (62076174).

\bibliography{anthology,custom}

\begin{thebibliography}{34}
\expandafter\ifx\csname natexlab\endcsname\relax\def\natexlab#1{#1}\fi

\bibitem[{Aji et~al.(2020)Aji, Bogoychev, Heafield, and
  Sennrich}]{aji-etal-2020-neural}
Alham~Fikri Aji, Nikolay Bogoychev, Kenneth Heafield, and Rico Sennrich. 2020.
\newblock \href {https://doi.org/10.18653/v1/2020.acl-main.688} {In neural
  machine translation, what does transfer learning transfer?}
\newblock In \emph{Proceedings of the 58th Annual Meeting of the Association
  for Computational Linguistics}, pages 7701--7710, Online. Association for
  Computational Linguistics.

\bibitem[{Artetxe et~al.(2017)Artetxe, Labaka, Agirre, and
  Cho}]{artetxe2017unsupervised}
Mikel Artetxe, Gorka Labaka, Eneko Agirre, and Kyunghyun Cho. 2017.
\newblock Unsupervised neural machine translation.
\newblock \emph{arXiv preprint arXiv:1710.11041}.

\bibitem[{Bojar et~al.(2017)Bojar, Chatterjee, Federmann, Graham, Haddow,
  Huang, Huck, Koehn, Liu, Logacheva, Monz, Negri, Post, Rubino, Specia, and
  Turchi}]{bojar-etal-2017-findings}
Ond{\v{r}}ej Bojar, Rajen Chatterjee, Christian Federmann, Yvette Graham, Barry
  Haddow, Shujian Huang, Matthias Huck, Philipp Koehn, Qun Liu, Varvara
  Logacheva, Christof Monz, Matteo Negri, Matt Post, Raphael Rubino, Lucia
  Specia, and Marco Turchi. 2017.
\newblock \href {https://doi.org/10.18653/v1/W17-4717} {Findings of the 2017
  conference on machine translation ({WMT}17)}.
\newblock In \emph{Proceedings of the Second Conference on Machine
  Translation}, pages 169--214, Copenhagen, Denmark. Association for
  Computational Linguistics.

\bibitem[{Cheng et~al.(2017)Cheng, Liu, Yang, Sun, and Xu}]{cheng2019joint}
Yong Cheng, Yang Liu, Qian Yang, Maosong Sun, and Wei Xu. 2017.
\newblock Joint training for pivot-based neural machine translation.
\newblock In \emph{Proceedings of the Twenty-Sixth International Joint
  Conference on Artificial Intelligence}, pages 3974--3980.

\bibitem[{Conneau et~al.(2018)Conneau, Lample, Ranzato, Denoyer, and
  J{\'e}gou}]{conneau2018word}
Alexis Conneau, Guillaume Lample, Marc'Aurelio Ranzato, Ludovic Denoyer, and
  Herv{\'e} J{\'e}gou. 2018.
\newblock Word translation without parallel data.
\newblock In \emph{Proceedings of the 6th International Conference on Learning
  Representations}.

\bibitem[{Dabre et~al.(2017)Dabre, Nakagawa, and
  Kazawa}]{dabre-etal-2017-empirical}
Raj Dabre, Tetsuji Nakagawa, and Hideto Kazawa. 2017.
\newblock \href {https://aclanthology.org/Y17-1038} {An empirical study of
  language relatedness for transfer learning in neural machine translation}.
\newblock In \emph{Proceedings of the 31st Pacific Asia Conference on Language,
  Information and Computation}, pages 282--286. The National University
  (Phillippines).

\bibitem[{Ding et~al.(2018)Ding, Utiyama, and Sumita}]{ding2018nova}
Chenchen Ding, Masao Utiyama, and Eiichiro Sumita. 2018.
\newblock Nova: A feasible and flexible annotation system for joint
  tokenization and part-of-speech tagging.
\newblock \emph{ACM Transactions on Asian and Low-Resource Language Information
  Processing (TALLIP)}, 18(2):1--18.

\bibitem[{Firat et~al.(2016)Firat, Cho, and Bengio}]{firat-etal-2016-multi}
Orhan Firat, Kyunghyun Cho, and Yoshua Bengio. 2016.
\newblock \href {https://doi.org/10.18653/v1/N16-1101} {Multi-way, multilingual
  neural machine translation with a shared attention mechanism}.
\newblock In \emph{Proceedings of the 2016 Conference of the North {A}merican
  Chapter of the Association for Computational Linguistics: Human Language
  Technologies}, pages 866--875, San Diego, California. Association for
  Computational Linguistics.

\bibitem[{Gheini and May(2019)}]{gheini2019universal}
Mozhdeh Gheini and Jonathan May. 2019.
\newblock A universal parent model for low-resource neural machine translation
  transfer.
\newblock \emph{arXiv preprint arXiv:1909.06516}.

\bibitem[{Gu et~al.(2018{\natexlab{a}})Gu, Hassan, Devlin, and
  Li}]{gu-etal-2018-universal}
Jiatao Gu, Hany Hassan, Jacob Devlin, and Victor~O.K. Li. 2018{\natexlab{a}}.
\newblock \href {https://doi.org/10.18653/v1/N18-1032} {Universal neural
  machine translation for extremely low resource languages}.
\newblock In \emph{Proceedings of the 2018 Conference of the North {A}merican
  Chapter of the Association for Computational Linguistics: Human Language
  Technologies, Volume 1 (Long Papers)}, pages 344--354, New Orleans,
  Louisiana. Association for Computational Linguistics.

\bibitem[{Gu et~al.(2018{\natexlab{b}})Gu, Wang, Chen, Li, and
  Cho}]{gu-etal-2018-meta}
Jiatao Gu, Yong Wang, Yun Chen, Victor O.~K. Li, and Kyunghyun Cho.
  2018{\natexlab{b}}.
\newblock \href {https://doi.org/10.18653/v1/D18-1398} {Meta-learning for
  low-resource neural machine translation}.
\newblock In \emph{Proceedings of the 2018 Conference on Empirical Methods in
  Natural Language Processing}, pages 3622--3631, Brussels, Belgium.
  Association for Computational Linguistics.

\bibitem[{Johnson et~al.(2017)Johnson, Schuster, Le, Krikun, Wu, Chen, Thorat,
  Vi{\'e}gas, Wattenberg, Corrado, Hughes, and
  Dean}]{johnson-etal-2017-googles}
Melvin Johnson, Mike Schuster, Quoc~V. Le, Maxim Krikun, Yonghui Wu, Zhifeng
  Chen, Nikhil Thorat, Fernanda Vi{\'e}gas, Martin Wattenberg, Greg Corrado,
  Macduff Hughes, and Jeffrey Dean. 2017.
\newblock \href {https://doi.org/10.1162/tacl_a_00065} {{G}oogle{'}s
  multilingual neural machine translation system: Enabling zero-shot
  translation}.
\newblock \emph{Transactions of the Association for Computational Linguistics},
  5:339--351.

\bibitem[{Kim et~al.(2019)Kim, Gao, and Ney}]{kim-etal-2019-effective}
Yunsu Kim, Yingbo Gao, and Hermann Ney. 2019.
\newblock \href {https://doi.org/10.18653/v1/P19-1120} {Effective cross-lingual
  transfer of neural machine translation models without shared vocabularies}.
\newblock In \emph{Proceedings of the 57th Annual Meeting of the Association
  for Computational Linguistics}, pages 1246--1257, Florence, Italy.
  Association for Computational Linguistics.

\bibitem[{Kim et~al.(2018)Kim, Geng, and Ney}]{kim-etal-2018-improving}
Yunsu Kim, Jiahui Geng, and Hermann Ney. 2018.
\newblock \href {https://doi.org/10.18653/v1/D18-1101} {Improving unsupervised
  word-by-word translation with language model and denoising autoencoder}.
\newblock In \emph{Proceedings of the 2018 Conference on Empirical Methods in
  Natural Language Processing}, pages 862--868, Brussels, Belgium. Association
  for Computational Linguistics.

\bibitem[{Kocmi and Bojar(2018)}]{kocmi-bojar-2018-trivial}
Tom Kocmi and Ond{\v{r}}ej Bojar. 2018.
\newblock \href {https://doi.org/10.18653/v1/W18-6325} {Trivial transfer
  learning for low-resource neural machine translation}.
\newblock In \emph{Proceedings of the Third Conference on Machine Translation:
  Research Papers}, pages 244--252, Brussels, Belgium. Association for
  Computational Linguistics.

\bibitem[{Koehn and Knowles(2017)}]{koehn-knowles-2017-six}
Philipp Koehn and Rebecca Knowles. 2017.
\newblock \href {https://doi.org/10.18653/v1/W17-3204} {Six challenges for
  neural machine translation}.
\newblock In \emph{Proceedings of the First Workshop on Neural Machine
  Translation}, pages 28--39, Vancouver. Association for Computational
  Linguistics.

\bibitem[{Kudo and Richardson(2018)}]{kudo-richardson-2018-sentencepiece}
Taku Kudo and John Richardson. 2018.
\newblock \href {https://doi.org/10.18653/v1/D18-2012} {{S}entence{P}iece: A
  simple and language independent subword tokenizer and detokenizer for neural
  text processing}.
\newblock In \emph{Proceedings of the 2018 Conference on Empirical Methods in
  Natural Language Processing: System Demonstrations}, pages 66--71, Brussels,
  Belgium. Association for Computational Linguistics.

\bibitem[{Lample et~al.(2018{\natexlab{a}})Lample, Conneau, Denoyer, and
  Ranzato}]{lample2018unsupervised}
Guillaume Lample, Alexis Conneau, Ludovic Denoyer, and Marc'Aurelio Ranzato.
  2018{\natexlab{a}}.
\newblock Unsupervised machine translation using monolingual corpora only.
\newblock In \emph{Proceedings of the 6th International Conference on Learning
  Representations}.

\bibitem[{Lample et~al.(2018{\natexlab{b}})Lample, Ott, Conneau, Denoyer, and
  Ranzato}]{lample-etal-2018-phrase}
Guillaume Lample, Myle Ott, Alexis Conneau, Ludovic Denoyer, and Marc{'}Aurelio
  Ranzato. 2018{\natexlab{b}}.
\newblock \href {https://doi.org/10.18653/v1/D18-1549} {Phrase-based {\&}
  neural unsupervised machine translation}.
\newblock In \emph{Proceedings of the 2018 Conference on Empirical Methods in
  Natural Language Processing}, pages 5039--5049, Brussels, Belgium.
  Association for Computational Linguistics.

\bibitem[{Lee et~al.(2017)Lee, Cho, and Hofmann}]{lee-etal-2017-fully}
Jason Lee, Kyunghyun Cho, and Thomas Hofmann. 2017.
\newblock \href {https://doi.org/10.1162/tacl_a_00067} {Fully character-level
  neural machine translation without explicit segmentation}.
\newblock \emph{Transactions of the Association for Computational Linguistics},
  5:365--378.

\bibitem[{Lin et~al.(2019)Lin, Chen, Lee, Li, Zhang, Xia, Rijhwani, He, Zhang,
  Ma, Anastasopoulos, Littell, and Neubig}]{lin-etal-2019-choosing}
Yu-Hsiang Lin, Chian-Yu Chen, Jean Lee, Zirui Li, Yuyan Zhang, Mengzhou Xia,
  Shruti Rijhwani, Junxian He, Zhisong Zhang, Xuezhe Ma, Antonios
  Anastasopoulos, Patrick Littell, and Graham Neubig. 2019.
\newblock \href {https://doi.org/10.18653/v1/P19-1301} {Choosing transfer
  languages for cross-lingual learning}.
\newblock In \emph{Proceedings of the 57th Annual Meeting of the Association
  for Computational Linguistics}, pages 3125--3135, Florence, Italy.
  Association for Computational Linguistics.

\bibitem[{Miceli~Barone(2016)}]{miceli-barone-2016-towards}
Antonio~Valerio Miceli~Barone. 2016.
\newblock \href {https://doi.org/10.18653/v1/W16-1614} {Towards cross-lingual
  distributed representations without parallel text trained with adversarial
  autoencoders}.
\newblock In \emph{Proceedings of the 1st Workshop on Representation Learning
  for {NLP}}, pages 121--126, Berlin, Germany. Association for Computational
  Linguistics.

\bibitem[{Nguyen and Chiang(2017)}]{nguyen-chiang-2017-transfer}
Toan~Q. Nguyen and David Chiang. 2017.
\newblock \href {https://aclanthology.org/I17-2050} {Transfer learning across
  low-resource, related languages for neural machine translation}.
\newblock In \emph{Proceedings of the Eighth International Joint Conference on
  Natural Language Processing (Volume 2: Short Papers)}, pages 296--301,
  Taipei, Taiwan. Asian Federation of Natural Language Processing.

\bibitem[{{\"O}stling et~al.(2016){\"O}stling, Tiedemann
  et~al.}]{ostling2016efficient}
Robert {\"O}stling, J{\"o}rg Tiedemann, et~al. 2016.
\newblock Efficient word alignment with markov chain monte carlo.
\newblock \emph{The Prague Bulletin of Mathematical Linguistics}.

\bibitem[{Post(2018)}]{post2018call}
Matt Post. 2018.
\newblock A call for clarity in reporting bleu scores.
\newblock In \emph{Proceedings of the Third Conference on Machine Translation:
  Research Papers}, pages 186--191.

\bibitem[{Sennrich et~al.(2016{\natexlab{a}})Sennrich, Haddow, and
  Birch}]{sennrich-etal-2016-improving}
Rico Sennrich, Barry Haddow, and Alexandra Birch. 2016{\natexlab{a}}.
\newblock \href {https://doi.org/10.18653/v1/P16-1009} {Improving neural
  machine translation models with monolingual data}.
\newblock In \emph{Proceedings of the 54th Annual Meeting of the Association
  for Computational Linguistics (Volume 1: Long Papers)}, pages 86--96, Berlin,
  Germany. Association for Computational Linguistics.

\bibitem[{Sennrich et~al.(2016{\natexlab{b}})Sennrich, Haddow, and
  Birch}]{sennrich-etal-2016-neural}
Rico Sennrich, Barry Haddow, and Alexandra Birch. 2016{\natexlab{b}}.
\newblock \href {https://doi.org/10.18653/v1/P16-1162} {Neural machine
  translation of rare words with subword units}.
\newblock In \emph{Proceedings of the 54th Annual Meeting of the Association
  for Computational Linguistics (Volume 1: Long Papers)}, pages 1715--1725,
  Berlin, Germany. Association for Computational Linguistics.

\bibitem[{S{\o}gaard et~al.(2018)S{\o}gaard, Ruder, and
  Vuli{\'c}}]{sogaard-etal-2018-limitations}
Anders S{\o}gaard, Sebastian Ruder, and Ivan Vuli{\'c}. 2018.
\newblock \href {https://doi.org/10.18653/v1/P18-1072} {On the limitations of
  unsupervised bilingual dictionary induction}.
\newblock In \emph{Proceedings of the 56th Annual Meeting of the Association
  for Computational Linguistics (Volume 1: Long Papers)}, pages 778--788,
  Melbourne, Australia. Association for Computational Linguistics.

\bibitem[{Tiedemann(2012)}]{tiedemann-2012-parallel}
J{\"o}rg Tiedemann. 2012.
\newblock \href
  {http://www.lrec-conf.org/proceedings/lrec2012/pdf/463_Paper.pdf} {Parallel
  data, tools and interfaces in {OPUS}}.
\newblock In \emph{Proceedings of the Eighth International Conference on
  Language Resources and Evaluation ({LREC}'12)}, pages 2214--2218, Istanbul,
  Turkey. European Language Resources Association (ELRA).

\bibitem[{Vaswani et~al.(2017)Vaswani, Shazeer, Parmar, Uszkoreit, Jones,
  Gomez, Kaiser, and Polosukhin}]{vaswani2017attention}
Ashish Vaswani, Noam Shazeer, Niki Parmar, Jakob Uszkoreit, Llion Jones,
  Aidan~N Gomez, {\L}ukasz Kaiser, and Illia Polosukhin. 2017.
\newblock Attention is all you need.
\newblock In \emph{Advances in neural information processing systems}, pages
  5998--6008.

\bibitem[{Vincent et~al.(2008)Vincent, Larochelle, Bengio, and
  Manzagol}]{vincentextracting}
Pascal Vincent, Hugo Larochelle, Yoshua Bengio, and Pierre-Antoine Manzagol.
  2008.
\newblock Extracting and composing robust features with denoising autoencoders.
\newblock In \emph{Proceedings of the 25th International Conference on Machine
  Learning}.

\bibitem[{Wolf et~al.(2020)Wolf, Debut, Sanh, Chaumond, Delangue, Moi, Cistac,
  Rault, Louf, Funtowicz, Davison, Shleifer, von Platen, Ma, Jernite, Plu, Xu,
  Le~Scao, Gugger, Drame, Lhoest, and Rush}]{wolf-etal-2020-transformers}
Thomas Wolf, Lysandre Debut, Victor Sanh, Julien Chaumond, Clement Delangue,
  Anthony Moi, Pierric Cistac, Tim Rault, Remi Louf, Morgan Funtowicz, Joe
  Davison, Sam Shleifer, Patrick von Platen, Clara Ma, Yacine Jernite, Julien
  Plu, Canwen Xu, Teven Le~Scao, Sylvain Gugger, Mariama Drame, Quentin Lhoest,
  and Alexander Rush. 2020.
\newblock \href {https://doi.org/10.18653/v1/2020.emnlp-demos.6} {Transformers:
  State-of-the-art natural language processing}.
\newblock In \emph{Proceedings of the 2020 Conference on Empirical Methods in
  Natural Language Processing: System Demonstrations}, pages 38--45, Online.
  Association for Computational Linguistics.

\bibitem[{Zhang et~al.(2017)Zhang, Liu, Luan, and
  Sun}]{zhang-etal-2017-adversarial}
Meng Zhang, Yang Liu, Huanbo Luan, and Maosong Sun. 2017.
\newblock \href {https://doi.org/10.18653/v1/P17-1179} {Adversarial training
  for unsupervised bilingual lexicon induction}.
\newblock In \emph{Proceedings of the 55th Annual Meeting of the Association
  for Computational Linguistics (Volume 1: Long Papers)}, pages 1959--1970,
  Vancouver, Canada. Association for Computational Linguistics.

\bibitem[{Zoph et~al.(2016)Zoph, Yuret, May, and
  Knight}]{zoph-etal-2016-transfer}
Barret Zoph, Deniz Yuret, Jonathan May, and Kevin Knight. 2016.
\newblock \href {https://doi.org/10.18653/v1/D16-1163} {Transfer learning for
  low-resource neural machine translation}.
\newblock In \emph{Proceedings of the 2016 Conference on Empirical Methods in
  Natural Language Processing}, pages 1568--1575, Austin, Texas. Association
  for Computational Linguistics.

\end{thebibliography}
\bibliographystyle{acl_natbib}

\end{document}